%% file: main.tex
\documentclass[runningsheads]{llncs}

\usepackage{amsmath}
\usepackage{svg}
\usepackage[T1]{fontenc}
\usepackage{graphicx}
\usepackage[utf8]{inputenc}
\usepackage{url}
\usepackage{graphicx}
\usepackage{authblk}
\usepackage{hhline}
\usepackage{lipsum}
\usepackage{algorithm}
\usepackage{algorithmic}
\usepackage{amssymb}
\usepackage[hidelinks]{hyperref}

\usepackage{float}
\usepackage{subcaption}
\usepackage{enumitem}
\usepackage{multirow}
\usepackage[nameinlink,capitalize]{cleveref}
\usepackage{booktabs}
\usepackage{svg}

\newcommand{\bftab}{\fontseries{b}\selectfont}
\begin{document}

\input{text/0abstract.tex}
\input{text/1introduction.tex}

\input{text/2related_works.tex}
\input{text/3methods.tex}
\input{text/4experiments.tex}
\input{text/5discussion.tex}

\newpage

\bibliography{bibboi}
\newpage
\input{text/6supplementary}
\end{document}

%% file: text/0abstract.tex
\title{Diffusion Based Ambiguous Image Segmentation}


\author{Jakob Lønborg Christensen\inst{1}\orcidID{0009-0001-1510-6081} \and Morten Rieger Hannemose\inst{1}\orcidID{0000-0002-9956-9226} \and Anders Bjorholm Dahl\inst{1}\orcidID{0000-0002-0068-8170} \and Vedrana Andersen Dahl\inst{1}\orcidID{0000-0001-6734-5570}}

\authorrunning{Christensen et al.}

\institute{The Technical University of Denmark, Department of Applied Mathematics and Computer Science (DTU Compute)} 

%
\maketitle

\begin{abstract}
Medical image segmentation often involves inherent uncertainty due to variations in expert annotations. Capturing this uncertainty is an important goal and previous works have used various generative image models for the purpose of representing the full distribution of plausible expert ground truths. In this work, we explore the design space of diffusion models for generative segmentation, investigating the impact of noise schedules, prediction types, and loss weightings. Notably, we find that making the noise schedule harder with input scaling significantly improves performance. We conclude that $x$- and $v$-prediction outperform $\epsilon$-prediction, likely because the diffusion process is in the discrete segmentation domain. Many loss weightings achieve similar performance as long as they give enough weight to the end of the diffusion process. We base our experiments on the LIDC-IDRI lung lesion dataset and obtain state-of-the-art (SOTA) performance. Additionally, we introduce a randomly cropped variant of the LIDC-IDRI dataset that is better suited for uncertainty in image segmentation. Our model also achieves SOTA in this harder setting. 

\keywords{Diffusion models  \and medical image segmentation \and uncertainty modeling.}
\end{abstract}

%% file: text/1introduction.tex
\section{Introduction}

Image segmentation has witnessed remarkable advancements driven by the rise of deep learning. The ability of convolutional neural networks (CNNs) and their variants to learn complex patterns has led to breakthroughs across diverse applications, from object detection to semantic segmentation~\cite{unet,long2015fully,yolo,ulku2022survey}. In particular, medical image segmentation has significantly benefited from these developments, enabling more accurate delineation of anatomical structures and pathological regions, thereby improving diagnostic workflows and treatment planning.

However, medical image segmentation remains an inherently ambiguous task. Delineating structures often depends on subtle, context-dependent features in imaging data, which can lead to disagreements among experts. This variability highlights the importance of obtaining multiple expert annotations to capture a more comprehensive understanding of the data. Incorporating this diversity in medical imaging pipelines is crucial for building systems that can handle such intrinsic uncertainties.

To address this, models must effectively capture the uncertainty inherent in medical imaging tasks. While Bayesian methods and ensemble networks have been explored for this purpose, they often fall short of capturing the full posterior distribution and its pixel covariances~\cite{bayesian_segnet,prob_unet}. Recent efforts have proposed generative approaches, such as Phi-Seg \cite{phi_seg} and the Probabilistic UNet \cite{prob_unet,gen_prob_unet,gen_prob_unet2}, which aim to model the distribution of plausible segmentations more comprehensively. The generative models take pixel-wise covariances into account, producing plausible segmentation maps which an expert might have produced. Simpler models often fail to capture these covariances, resulting in blurry segmentations and nonsensical delineations.

Recently, diffusion models have emerged as a powerful family of generative models, achieving state-of-the-art (SOTA) results in various domains such as image synthesis and super-resolution \cite{ddpm,iDDPM,VDM,simple_diff2}. In this work, we aim to explore the potential of diffusion models for generative medical image segmentation. By comparing their performance with established methods such as Phi-Seg~\cite{phi_seg} and the Probabilistic UNet~\cite{prob_unet,gen_prob_unet}, we seek to determine their suitability for capturing the inherent uncertainties in medical image segmentation tasks. Our main contributions are:
\begin{enumerate}
    \item We investigate which variety of diffusion model is best suited for ambiguous image segmentation, based on recent diffusion research.
    \item Our best model improves the SOTA on the LIDC-IDRI dataset. 
    \item We introduce a randomly cropped variant of the LIDC-IDRI dataset which more faithfully captures uncertainty in segmentation and also achieve SOTA on this task.
\end{enumerate}

\begin{figure}
    \centering
    \includegraphics[width=1.0\linewidth]{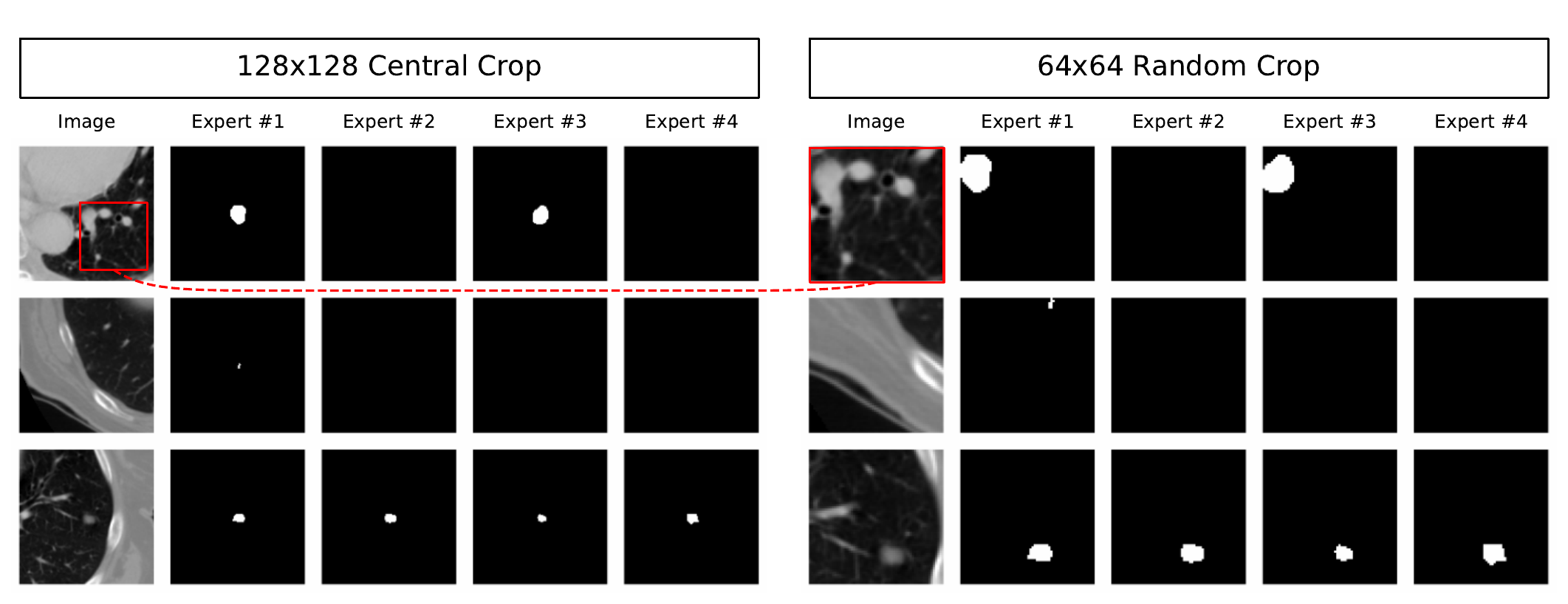}
    \caption{Samples from the LIDC-IDRI lung lesion segmentation dataset, where each image has 4 expert annotations. The same 3 images are shown in a central $128\times128$ crop and a random $64\times64$ crop. The red square indicates the first of these crops.}
    \label{fig:data_samples}
\end{figure}

%% file: text/2related_works.tex
\section{Background}
The Lung Image Database Consortium image collection (LIDC-IDRI) is a dataset commonly used for ambiguous medical image segmentation (see \cref{fig:data_samples}). Originally a 3D dataset consisting of 1018 thoracic CT images, it is usually used as 2D slices cropped centrally around the lesions in a $128 \times 128$ resolution. This yields 15096 2D images, which were separated into a 60-20-20 holdout split on a per patient basis. The unique feature of the dataset is that it contains 4 independent expert annotations for each lesion, making it ideal for modeling uncertainty. All images have at least one positive (non-empty) expert annotation.

The LIDC dataset, see \cref{fig:data_samples}, leaves little ambiguity regarding which structure in the image is the potential lesion, because of the central crop. Segmenting the lesion is relatively easy, and the harder part of the task is to predict how likely the ground truth is to be negative. Therefore, the segmentation task is more akin to simple classification plus a deterministic mask prediction. This motivates our proposal for a new variant of the dataset: randomly cropping a $64\times64$ subset of the images, distributed uniformly.

In order to compare masks, we use the dice coefficient given by $2\text{TP}/(2\text{TP}+\text{FP}+\text{FN})$ and the intersection over union given by $\text{TP}/(\text{TP}+\text{FP}+\text{FN})$. 
If the denominator is 0 in either dice or IoU, the value is instead 1 as it represents both an empty prediction and ground truth.
The most significant metric for ambiguous segmentation is the generalized energy distance (GED), since it captures to which degree the ground truth and learned distributions match\cite{phi_seg,prob_unet}. It is defined by $D^2_\text{GED}(\mathcal{S},\mathcal{Y})=2 \mathbb{E}[d(s,y)]-\mathbb{E}[d(s,s')]-\mathbb{E}[d(y,y')]$, where the distance metric is $d(\cdot,\cdot)=1-IoU(\cdot,\cdot)$.
The variables $s$ and $s'$ are independent samples from the learned distribution $\mathcal{S}$,  and $y$ and $y'$ are independent samples from the ground truth distribution ($\mathcal{Y}$). The GED score always falls in the 0 to 2 interval, where lower is better. In practice, the expectations are computed by Monte Carlo simulation.

Phi-Seg~\cite{phi_seg} was one of the first works to model the uncertainty of medical image segmentation with a generative model, which outperformed network ensembles and approximate Bayesian methods. They used a hierarchical variational autoencoder (VAE), with a latent variable at each resolution. The models were mainly evaluated on data from the LIDC Lung Lesion dataset. The authors also published their preprocessed data, making comparison between methods much easier for the field. This preprocessed data was also used in our work.

The Probabilistic UNet~\cite{prob_unet} used a very different generative model compared to Phi-Seg. They split the model into three networks; A segmentation UNet, a prior network, and posterior network. The role of both the posterior and prior networks are to encode the input image into a latent space, representing the space of possible segmentations. The posterior network is given the ground truth, while the prior is not. Both the prior and posterior networks output the mean and variance of the latent space. The latent space is an axis-aligned Gaussian distribution (AA), meaning the covariance matrix is diagonal. The parameters of these networks were optimized with the KL-divergence between the prior and posterior distributions, forcing them to be close to each other. During training, the posterior latent distribution is used to sample a distribution of masks. Of course, this means the network uses the ground truth during training. However the trick of the paper is to use the prior latent distribution during inference, hoping the latent distributions are close enough to each other that the shift in distribution is not too large. 

Later, the Probabilistic UNet was expanded upon with the Generalized Probabilistic UNet~\cite{gen_prob_unet,gen_prob_unet2}, by using a selection of different latent distributions. Specifically, they found success with a mixture of axis-aligned Gaussians (AA mix), a full covariance Gaussian (FC) and a mixture of full covariance Gaussians (FC mix). For replication we use the optimal hyperparameters as reported \cite{gen_prob_unet}, except for the standard Probabilistic UNet (aka AA) where we found models with parameter $\beta=10$ better than $\beta=1$.

There are a few downsides to the Probabilistic UNet model formulations. Many parameters are spent on the prior and posterior networks with a similar size as the segmentation network. The posterior network is not used during inference and we are therefore not using these parameters to segment directly. Additionally, the training procedure is relatively unstable.


Others have explored using diffusion models for segmentation~\cite{seg_diff,medsegdiff,medsegdiff2,ensemble_diff,amb_diff_seg}. Notably, Rahman et al.~\cite{amb_diff_seg} used a diffusion model to attack the same problem as us, however they based the structure of the networks on the Probabilistic UNet~\cite{prob_unet}. They used a prior and posterior network to encode the input image into a latent space. The output of this latent space was not used anywhere else in the model\footnote{The \href{https://github.com/aimansnigdha/Ambiguous-Medical-Image-Segmentation-using-Diffusion-Models/blob/a9677afe4eedb163db478cd01ea8228161448b33/guided_diffusion/gaussian_diffusion.py\#L977C34-L977C39}{official code on GitHub} was used to confirm this}, allowing the KL-terms to be trivially satisfied by e.g.\ always predicting the same distribution regardless of the sample. 
We were unable to replicate their results, which prevented a fair comparison.

%% file: text/3methods.tex
\section{Methods}

\subsection{Diffusion Model}
We use a continuous time diffusion model ranging from time $t=0$ (data) to $t=1$ (noise). The noisy diffusion sample, $\mathbf{x}_t$, at time $t$ is given by 
\begin{equation} \label{eq:latent_diff}
    \mathbf{x}_t = \alpha(t) \mathbf{x}_0 + \sigma(t) \mathbf{\epsilon},
\end{equation}
where $\mathbf{x}_0$ is data and $\mathbf{\epsilon}$ is i.i.d unit Gaussian noise. In our case the data is the mask for a given image. The model is conditioned on the image by simple concatenation with the noisy mask in the channel dimension. $\alpha(t)$ and $\sigma(t)$ parameterize the noise schedule in such a way that the diffusion sample is pure data at $t=0$ and pure noise at $t=1$. Adding noise to the data is easy and is known as the forward process. Denoising to obtain the original data (aka the reverse process) is a much harder problem. If one can model the reverse process accurately, it will yield a generative model.

We employ a convolutional neural network (CNN) to predict the mean of the conditional distribution $p(\mathbf{x}_0 | \mathbf{x}_t)$, i.e. predicting the data, from a noisy latent sample. Using this prediction, \cite{VDM} showed that the posterior distribution for an arbitrary timestep $p(\mathbf{x}_s | \mathbf{x}_t)$ is given by
\begin{equation}
    \mathbf{x}_s = \sqrt{\alpha(s)^2 / \alpha(t)^2} ((1-c) \mathbf{x}_t -c \alpha(t) \mathbf{x}_0) + \sqrt{c(1-\alpha(s)^2)} \epsilon,
\end{equation}
where $\epsilon \sim \mathcal{N}(0,1)$ and $c = 1- \alpha(s)^2\sigma(s)^{-2}\alpha(t)^{-2}\sigma(t)^2$.
This equation allows us to do ancestral sampling. Starting with a sample from the noise distribution at $t=1$, one can sample the previous timesteps by using the equation above. 

\subsection{Noise Schedule and Input Scaling}
The noise schedule is parameterized by $\gamma: \left[0,1\right] \rightarrow \left[0,1\right]$, which is a monotonically decreasing function of time. We use a variance preserving noise schedule, where the coefficients are given by
\begin{equation}
    \alpha(t) = \sqrt{\gamma(t)}, \quad \sigma(t) = \sqrt{1 - \gamma(t)}.
\end{equation}
The variance preserving property ($\alpha^2+\sigma^2=1$) enables parameterization of both sets of coefficients with a single function. A common choice for the noise schedule which our model will utilize is the cosine schedule given by
\begin{equation}
\gamma(t) = \cos\left( \frac{t  \pi}{2} \right)^2.
\end{equation}

\begin{figure}
    \centering
    \includegraphics[width=0.95\textwidth]{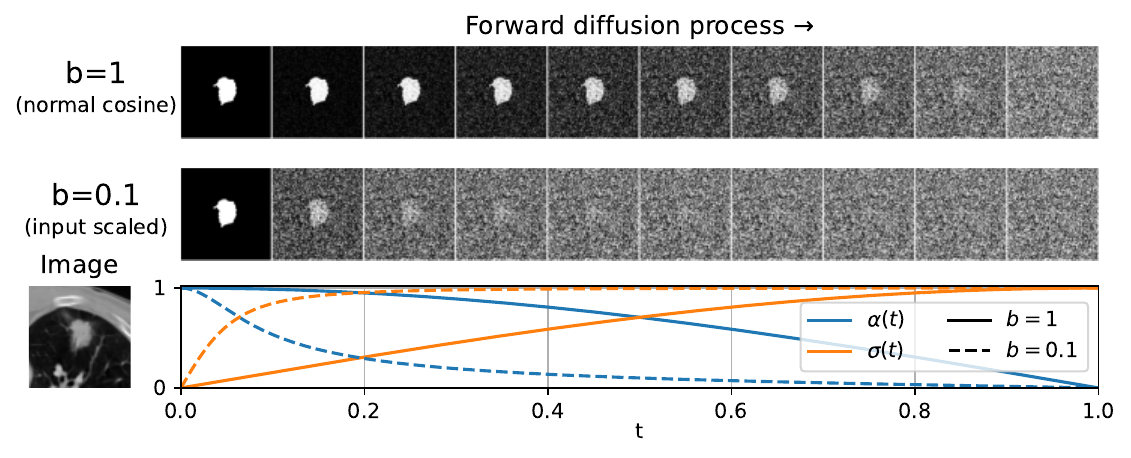}
    \caption{The cosine noise schedule with latent diffusion samples $x_t$ for linearly spaced values of $t$ from 0 to 1.}
    \label{fig:ignore_image_problem}
\end{figure}

Consider the top row of latent samples in \cref{fig:ignore_image_problem} generated from the cosine schedule. Reconstructing the segmentation is quite easy, even with a relatively high noise level. The fact that the mask is binary and highly correlated with nearby pixels can make the reconstruction task trivial. When $t$ is less than $\sim0.7$, you can reconstruct the mask without considering the image. This would attain a small training loss but resulting in a poor model that ignores the image during inference. 

Input scaling can be used to make diffusion noise schedules harder~\cite{input_scaling}. Input scaling was originally introduced to deal with large images since increasing the number of pixels decreases the effect of uncorrelated pixel noise. The idea behind input scaling is to make the noise schedule harder by lowering the signal-to-noise ratio (SNR). The SNR is given by
\begin{equation}
\label{eq:snr}
    \text{SNR}(t)=\frac{\alpha(t)^2}{\sigma(t)^2}=\frac{{\gamma(t)}}{1-\gamma(t)}.
\end{equation}
The SNR is lowered by multiplying the data (or input), $\alpha(t)$, with some constant $b \in [0,1]$, called the input scale. It turns out, we can find a perturbed expression for the noise schedule $\gamma_b(t)$ which has the same effect on the SNR, while maintaining unit variance. This noise schedule must satisfy the equation
\begin{equation}
    \frac{\gamma_b(t)}{1-\gamma_b(t)} = \frac{b^2 \gamma(t)}{1-\gamma(t)},
\end{equation}
and isolating $\gamma_b(t)$ yields the expression 
\begin{equation}
    \gamma_b(t)=\frac{b^2 \gamma(t)}{(b^2-1) \gamma(t) + 1}.
\end{equation}
Thus, all equations involving the noise schedule can be reused, by replacing $\gamma(t)$ with the input scaled schedule $\gamma_b(t)$. 

\subsection{Prediction Type}
The goal of the diffusion network is to predict the data given a noisy latent sample. The prediction can be done in many ways. Two common options are either to predict the data ($\mathbf{x}_0$) directly or to predict the noise ($\epsilon$). Each of these predictions parameterize the other, based on \cref{eq:latent_diff} since these are the only two unknown variables in the equation (from the perspective of the network). We also consider a third option known as $v$-prediction~\cite{prog_distil}, where $\mathbf{v}=\alpha(t) \epsilon - \sigma(t) \mathbf{x}_0$. This variable also parameterizes $\mathbf{x}_0$ and $\epsilon$.

\subsection{Loss Weighting}
\begin{figure}[H]
    \centering
    \includegraphics[width=0.90\textwidth]{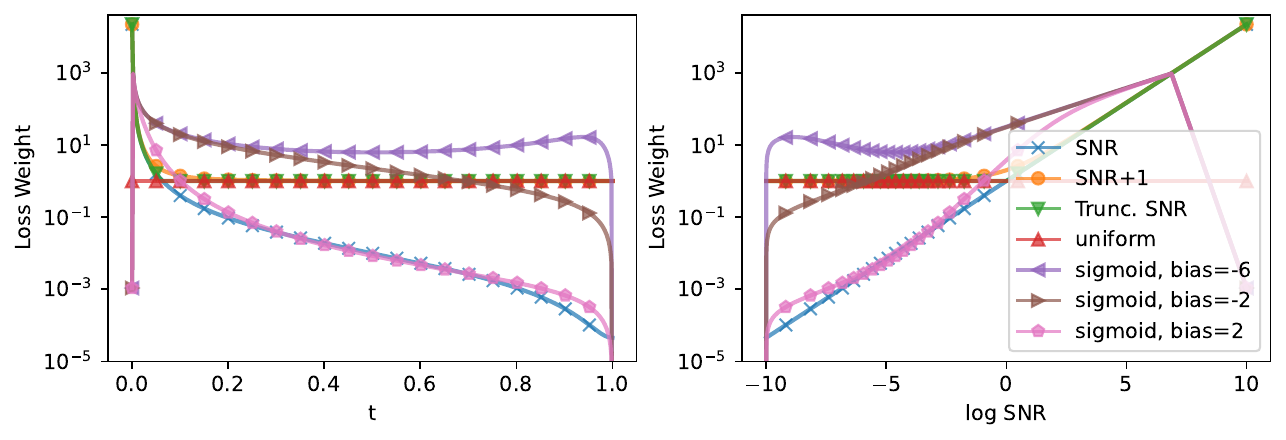}
    \caption{A visualization of various loss weightings. The cosine noise schedule with input scale $b=0.1$ was used.}
    \label{fig:loss_weights_vis}
\end{figure}

It was shown in \cite{VDM} that the evidence based lower bound (ELBO) for continuous time diffusion models can be written as
\begin{equation}
    L(\mathbf{x}) = \mathbb{E}_{t \sim \mathcal{U}(0, 1)} \left[ w(t) \|\mathbf{x}_0 - \hat{\mathbf{x}}\|^2 \right],
\end{equation}
where $\hat{\mathbf{x}}=\hat{\mathbf{x}}_\theta(\mathbf{x}_t, t)$ is the neural network prediction of the data, $\mathbf{x}_0$. The function $w(t)$ is a weighting function that can be used to emphasize learning at certain timesteps. A few common choices~\cite{prog_distil,simple_diff}, and the ones we use are
\begin{itemize}
    \item The signal-to-noise ratio, $w(t)=\text{SNR}(t)$, (\cref{eq:snr}).
    \item The truncated signal to noise ratio, $w(t)=\text{max}(\text{SNR}(t),1)$.
    \item The signal-to-noise ratio plus 1, $w(t)=\text{SNR}(t)+1$.
    \item The simple uniform weighting, $w(t)=1$.
    \item The Sigmoid weighting, given by $w(t)=-\frac{\text{d} \log\text{SNR}(t)}{\text{d}t} \tilde{\sigma}(\log{\text{SNR}(t)+\tilde{b}})$, where $\tilde{\sigma}(t)=(1+e^{-t})^{-1}$ is the sigmoid function, and $\tilde{b}$ is a bias parameter.
\end{itemize}
Various loss weightings are visualized in \cref{fig:loss_weights_vis}. The sigmoid weighting with $\tilde{b}=2$ is similar to SNR and smaller bias weights give more influence to the latter part of the diffusion process.

\subsection{Model Architecture}
The network architecture is a UNet~\cite{unet}. Our implementation is based on a modern version of a diffusion UNet from~\cite{iDDPM} including common additions such as residual connections, group ($k=32$) normalization, SiLU activations and attention layers after convolutions on the most downscaled blocks. Additionally, recent research concerning image generation with diffusion UNets have shown some architectural improvements that we have incorporated into our model. Specifically, Simple Diffusion~\cite{simple_diff,simple_diff2} showed using more residual blocks on downscaled parts of the UNet was beneficial.


%% file: text/4experiments.tex
\section{Experiments}
\begin{figure}[h]
    \centering
    \includegraphics[width=1.0\linewidth]{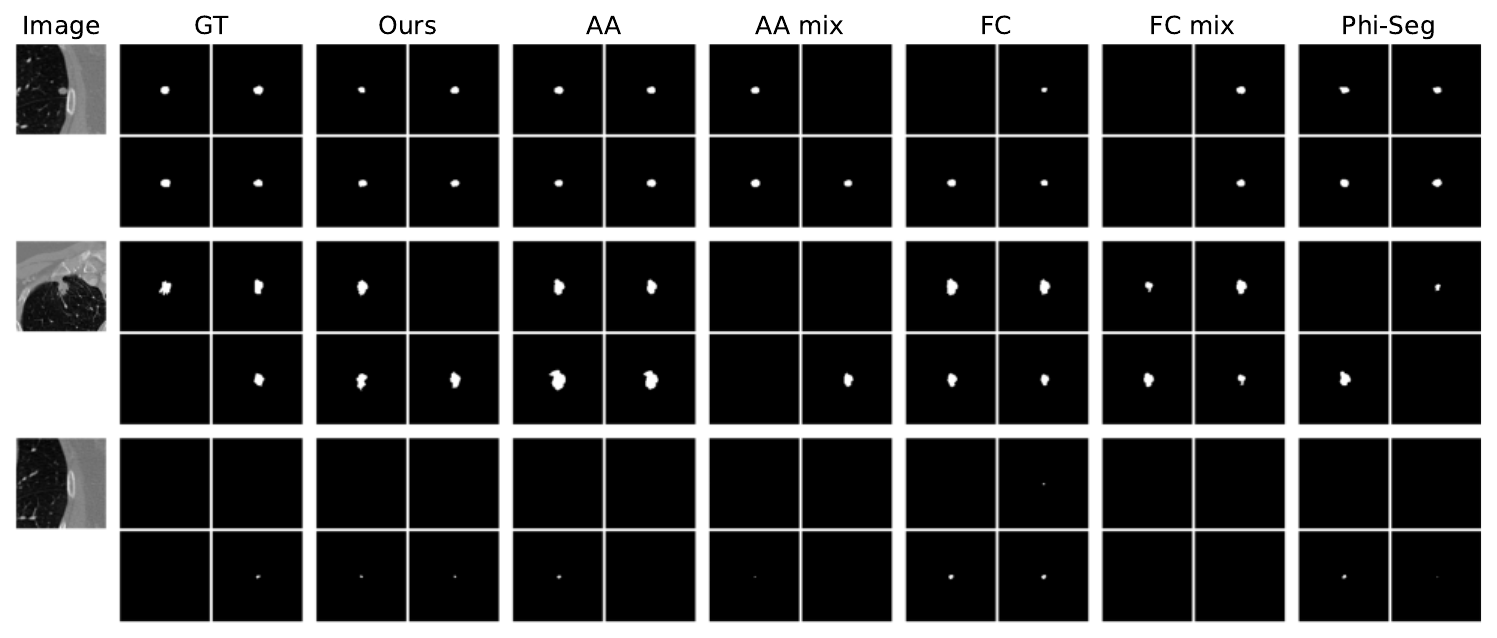}
    \caption{Results on samples from the test set for each model with $128\times128$ centrally cropped images.}
    \label{fig:comparison_qual}
\end{figure}

Our model is trained with the truncated SNR loss weighting and $x$-prediction. The cosine noise schedule was used, with an input scaling parameter with value $b=0.1$. A comparison of qualitative samples for our $128\times128$ centrally cropped models is shown in \cref{fig:comparison_qual} and quantitatively with mean metric values in \cref{tab:sota_performance_crop128}. We managed to improve performance for 4 predictions when applying post-processing (labeled pp) to the model outputs. The details of the postprocessing is described in the supplementary materials.

\begin{figure}[h]
    \centering
    \includegraphics[width=1.0\linewidth]{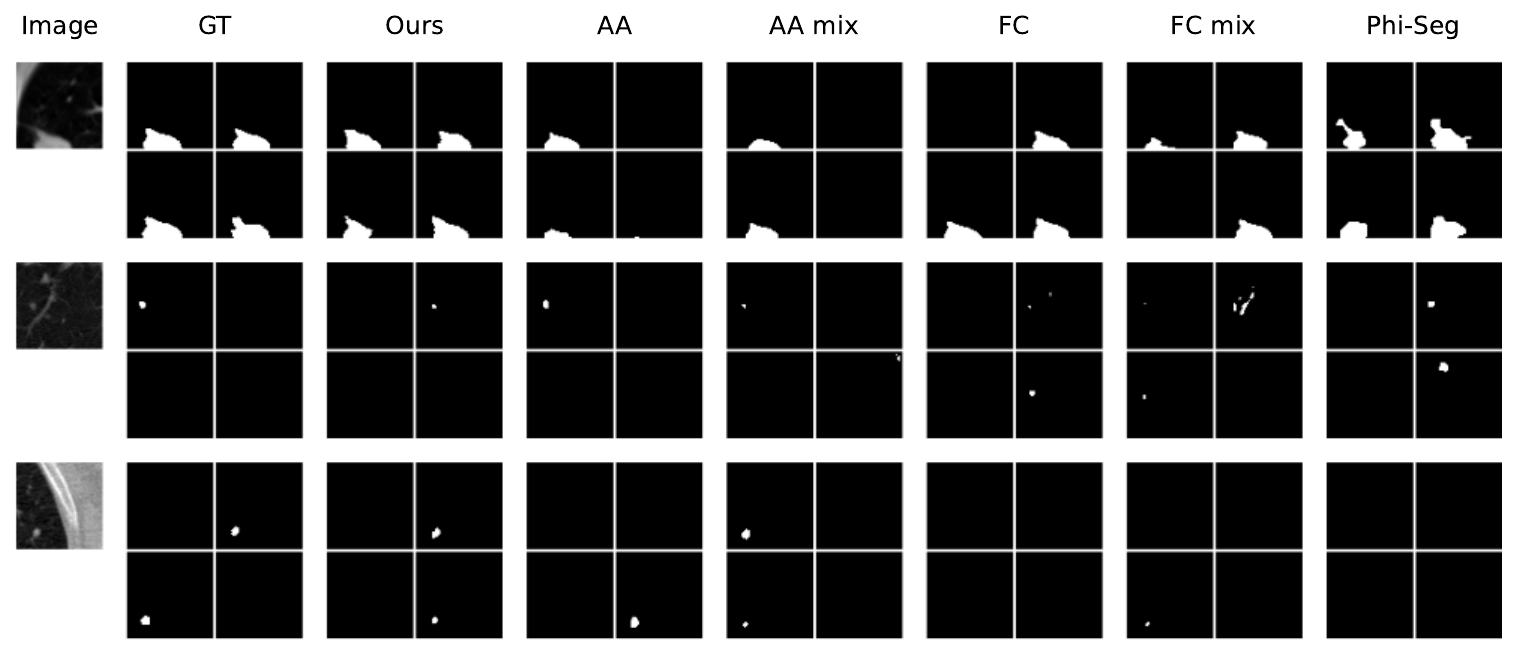}
    \caption{Results on samples from the test set for each model with $64\times64$ randomly cropped images.}
    \label{fig:comparison_qual_aug64}
\end{figure}

Our model is best by a slight margin in almost all metrics. When we compare the $64\times64$ randomly cropped models, our model is also best, but by a larger margin (see \cref{tab:sota_performance_crop64}). It is evident from \cref{fig:comparison_qual} the main aspect of variation between the models is the number of empty predictions. Comparatively,
when using a random crop (see \cref{fig:comparison_qual_aug64}), we observe more variation in not just lesion location but also shape.

We find that the second best model for the central crop and random crops is the standard Prob. UNet (AA) and the full covariance Gen. Prob. UNet (FC) respectively. The random crop task is arguably harder, perhaps explaining why the more complex latent distribution of FC is better for this task.

\begin{table}[h]
    \caption{Performance on the test set compared with the best current methods, on $128\times128$ centrally cropped images. The best performance is indicated in bold.}
    \label{tab:sota_performance_crop128}
    \centering
\begin{tabular}{l|p{1cm}p{1cm}p{1cm}p{1cm}|p{1cm}p{1cm}|l}
 & \multicolumn{4}{c|}{4 preds} & \multicolumn{2}{c|}{16 preds} &  \\ \hline
\textbf{Model} & \textbf{GED} & \textbf{\begin{tabular}[c]{@{}l@{}}GED \\ (pp)\end{tabular}} & \textbf{dice} & \textbf{\begin{tabular}[c]{@{}l@{}}dice \\ (pp)\end{tabular}} & \textbf{GED} & \textbf{dice} & \textbf{\begin{tabular}[c]{@{}l@{}}No. of\\ Params\end{tabular}} \\ \hline
Phi-Seg & 0.347 & 0.331 & 0.485 & 0.494 & 0.252 & 0.481 & 7322210 \\
Prob. UNet (AA) & 0.334 & 0.319 & {\bftab{0.488}} & 0.498 & 0.235 & {\bftab{0.491}} & 6847129 \\
Gen. Prob. UNet (AA mix) & 0.338 & 0.333 & 0.485 & 0.491 & 0.240 & 0.485 & 6864027 \\
Gen. Prob. UNet (FC) & 0.365 & 0.357 & 0.453 & 0.461 & 0.253 & 0.456 & 6844945 \\
Gen. Prob. UNet (FC mix) & 0.344 & 0.337 & 0.486 & 0.493 & 0.249 & 0.486 & 6882467 \\ \hline
Ours & {\bftab{0.330}} & {\bftab{0.304}} & {\bftab{0.488}} & {\bftab{0.507}} & {\bftab{0.233}} & 0.486 & 6891905
\end{tabular}
\end{table}

\begin{table}[h]
    \caption{Performances on the test set compared with the best current methods, on $64\times64$ randomly cropped images. The best performance is indicated in bold.}
    \label{tab:sota_performance_crop64}
    \centering
\begin{tabular}{l|p{1cm}p{1cm}p{1cm}p{1cm}|p{1cm}p{1cm}|l}
 & \multicolumn{4}{c|}{4 preds} & \multicolumn{2}{c|}{16 preds} &  \\ \hline
\textbf{Model} & \textbf{GED} & \textbf{\begin{tabular}[c]{@{}l@{}}GED \\ (pp)\end{tabular}} & \textbf{dice} & \textbf{\begin{tabular}[c]{@{}l@{}}dice \\ (pp)\end{tabular}} & \textbf{GED} & \textbf{dice} & \textbf{\begin{tabular}[c]{@{}l@{}}No. of\\ Params\end{tabular}} \\ \hline
Phi-Seg & 0.470 & 0.534 & 0.418 & 0.429 & 0.456 & 0.425 & 7322210 \\
Prob. UNet (AA) & 0.444 & 0.453 & 0.421 & 0.429 & 0.345 & 0.422 & 6847129 \\
Gen. Prob. UNet (AA mix) & 0.440 & 0.433 & 0.428 & 0.441 & 0.341 & 0.427 & 6864027 \\
Gen. Prob. UNet (FC) & 0.424 & 0.427 & 0.427 & 0.439 & 0.320 & 0.429 & 6844945 \\
Gen. Prob. UNet (FC mix) & 0.443 & 0.441 & 0.426 & 0.437 & 0.342 & 0.426 & 6882467 \\ \hline
Ours & {\bftab{0.419}} & {\bftab{0.405}} & {\bftab{0.438}} & {\bftab{0.459}} & {\bftab{0.312}} & {\bftab{0.442}} & 6891905
\end{tabular}
\end{table}

\begin{figure}[h]
    \centering
    \includegraphics[width=1.0\textwidth]{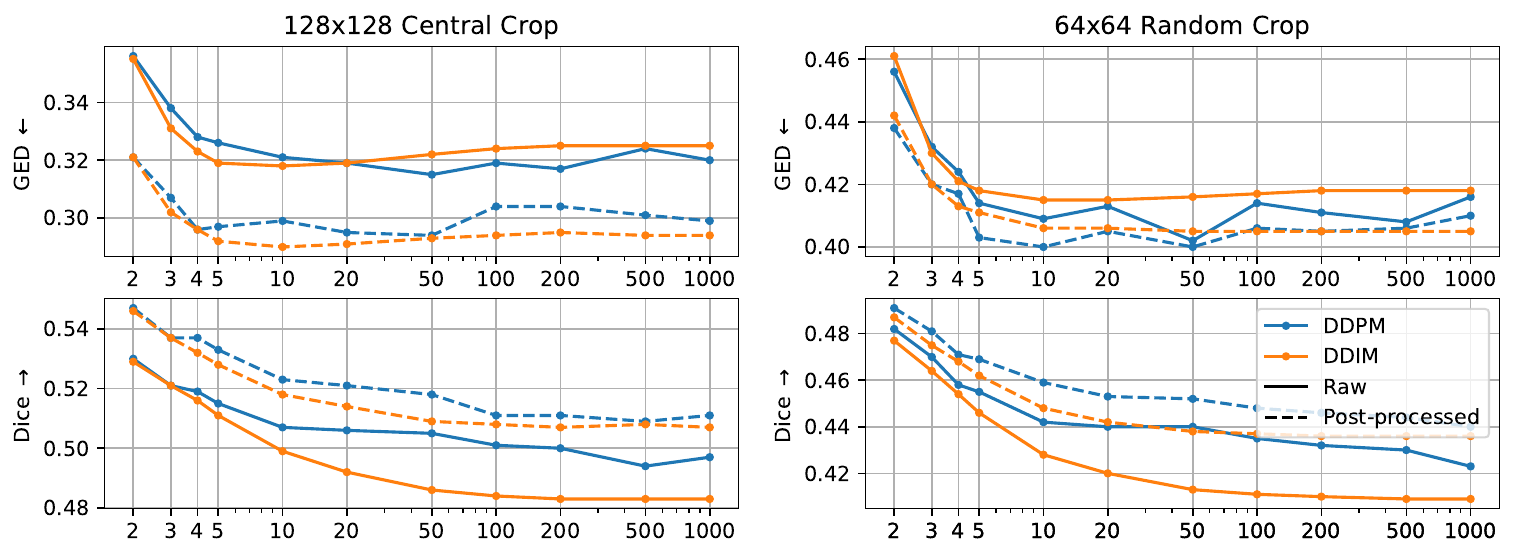}
    \caption{The effect of the number of timesteps used during inference on the GED and dice metrics. We used 4 predictions on images from the validation set.}
    \label{fig:timesteps}
\end{figure}

To find the optimal inference setup, we vary the number of timesteps as seen in \cref{fig:timesteps}. We try both the original DDPM~\cite{ddpm} sampling algorithm and the commonly used DDIM~\cite{DDIM} sampler. 
The performance curves in \cref{fig:timesteps} are somewhat surprising, as the best dice score is obtained with fewer timesteps. This might be explained by the fact that the network behaves more like a standard UNet when few function evaluations are used instead of the stochastic generative diffusion model. Based on these results, we use a 10 timesteps DDIM sampler when evaluating our $128\times128$ central crop models, and a 10 step DDPM sampler when evaluating our $64\times64$ random crop models.

\subsection{Ablation studies}

We trained models with a range of prediction types and loss weightings (see \cref{tab:pred_lw_comparison}). Models with $\epsilon$-prediction are consistently worse than their counterparts. Samples from $\epsilon$-prediction models retain significant Gaussian noise after the reverse diffusion process. Models with the SNR loss weighting are consistently worse than their counterparts for $x$ and $v$-prediction. The other three loss weightings are similar in performance, and we attribute most of their differences to random chance. Note that these models were only trained for 50k, compared to the 400k steps used for our final models.

\begin{figure}[h]
    \centering
    \includegraphics[width=0.9\linewidth]{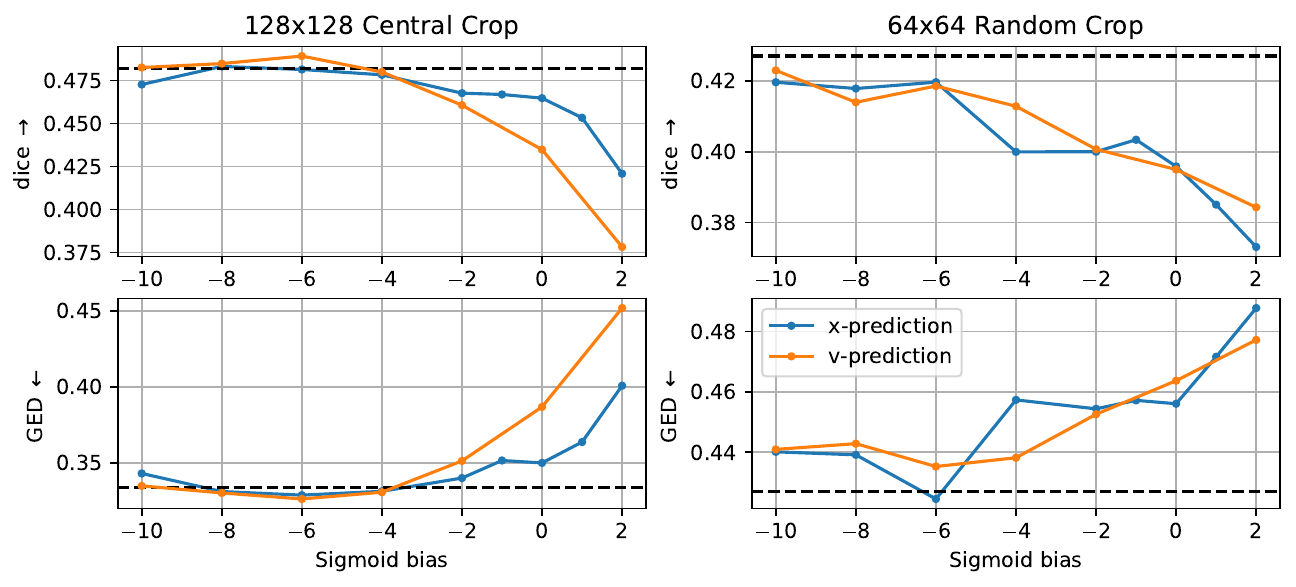}
    \caption{Performance measured in GED and dice for the sigmoid schedule with different biases. The dashed lines are the best scores (bold numbers) from \cref{tab:pred_lw_comparison}.}
    \label{fig:sigma_bias_sweep}
\end{figure}

We also trained models with the newer sigmoid loss\cite{VDM_pp,simple_diff2}, varying its bias parameter (see \cref{fig:sigma_bias_sweep}). For our tasks, a bias parameter of around $-6$ seems optimal and similar in performance to the best models in \cref{tab:pred_lw_comparison}.

\begin{table}[h]
\caption{Performance with different loss weightings and prediction targets. The number before the slash is GED and after is dice. Bold numbers indicate the best performance for each crop.}
\label{tab:pred_lw_comparison}
\centering
\begin{tabular}{c|c|p{2cm}p{2cm}p{2cm}p{2cm}}
\multirow{2}{*}{\textbf{Crop}} & \multirow{2}{*}{\textbf{\begin{tabular}[c]{@{}c@{}}Pred. \\ Type\end{tabular}}} & \multicolumn{4}{c}{\textbf{Loss Weighting}} \\
 &  & \multicolumn{1}{c}{\textbf{SNR}} & \multicolumn{1}{c}{\textbf{Trunc. SNR}} & \multicolumn{1}{c}{ \textbf{SNR+1}} & \multicolumn{1}{c}{\textbf{Uniform}} \\ \hline
\multirow{3}{*}{\begin{tabular}[c]{@{}c@{}}$128\times128$ \\ central \\ crop\end{tabular}} & \textbf{x} & 0.456/0.388 & 0.342/0.475 & 0.338/0.474 & 0.339/0.476  \\
 & \textbf{$\epsilon$} & 0.822/0.144 & 1.135/0.011 & 0.759/0.141 & 0.849/0.144  \\
 & \textbf{v} & 0.410/0.425 & 0.335/{\bftab{0.482}} & {\bftab{0.334}}/{\bftab{0.482}} & 0.337/0.479  \\ \hline
\multirow{3}{*}{\begin{tabular}[c]{@{}c@{}}$64\times64$ \\ random \\ Crop\end{tabular}} & \textbf{x} & 0.490/0.351 & 0.442/0.405 & 0.441/0.401 & 0.443/0.413  \\
 & \textbf{$\epsilon$} & 0.768/0.163 & 0.855/0.154 & 0.866/0.153 & 0.797/0.150  \\
 & \textbf{v} & 0.470/0.391 & 0.452/0.396 & 0.440/0.413 & {\bftab{0.427}}/{\bftab{0.427}} 
\end{tabular}
\end{table}

\begin{figure}[h]
    \centering
    \includegraphics[width=0.8\textwidth]{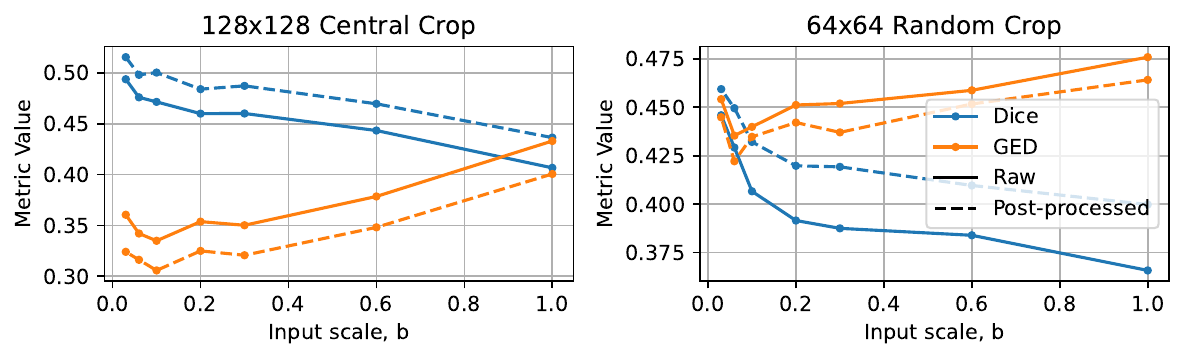}
    \caption{The effect of training models with different input scales ($b$) on the GED and dice metrics with 4 predictions per image. Curves are shown for raw and post-processed (pp) metrics on the validation set.}
    \label{fig:input_scale}
\end{figure}

We varied the bias parameter of input scaling, $b$ (see \cref{fig:input_scale}). Smaller values of $b$ yields a harder (noisier) noise schedule, while a value of $b=1$ is the same as using no input scaling. We try a range of options between 0 and 1 and find that $b=0.1$ is close to the optimum.

\subsection{Architecture Comparison}
The works used for comparison had different architectural details in their UNets. To account for this, we trained our models twice; once with the UNet architecture of Gen. Prob. UNet, and once with the diffusion UNet architecture. 

The number of parameters had to be reduced for the Prob. UNet methods, as a portion of the $\sim 7$ million total parameters are spent on their prior and posterior networks. This was done by scaling the number of channels but keeping other architectural details the same. The diffusion timestep embedding was removed when training a Gen. Prob. model on a diffusion UNet. Conversely, it was added when training a diffusion model on the Gen. Prob. UNet to provide timestep information. 
The results in \cref{tab:diff_vs_non_diff_unet}, show that the choice of UNet makes little difference in most cases. The standard Prob. UNet (AA) shows some preference for its native architecture.

\begin{table}[h]
\caption{Comparison of the two best UNet architectures from \cref{tab:sota_performance_crop128,tab:sota_performance_crop64}. They were measured with GED and dice metrics on the validation set. The best performance is indicated in bold.}
\label{tab:diff_vs_non_diff_unet}
\centering
\begin{tabular}{c|l|p{1.4cm}p{1.4cm}p{1.4cm}p{1.4cm}p{1.4cm}}
\multicolumn{1}{l|}{} & Architecture & AA & AA mix & FC & FC mix & Ours \\ \hline
\multirow{2}{*}{GED$\downarrow$} & Gen. Prob. UNet & \bftab{0.310} & \bftab{0.323} & 0.349 & 0.333 & \bftab{0.314}\\
 & Diff Unet & 0.329 & 0.337 & \bftab{0.338} & \bftab{0.322} & 0.318\\ \hline
\multirow{2}{*}{dice$\uparrow$} & Gen. Prob. UNet & \bftab{0.505} & \bftab{0.495} & 0.462 & 0.494 & 0.499\\
 & Diff Unet & 0.496 & 0.485 & \bftab{0.470} & \bftab{0.498} & 0.499
\end{tabular}
\end{table}

%% file: text/5discussion.tex
\section{Discussion}

Our model consistently outperformed the state-of-the-art, however the margins were relatively small. We theorize that the LIDC segmentation task was relatively well optimized already. This is supported by our observation of a larger performance gap when using the harder randomly cropped version of the task. Our model seemed agnostic to the task at hand, while different Gen. Prob. UNets were suited for either central or random crops. Another upside of our model is that the training loss is much more stable as there are no trade-off terms in the loss. The loss formulations for Phi-Seg and Prob. UNets are VAE-inspired and thus include different terms that need to be balanced. Training was comparatively more stable for the diffusion model, though there was one aspect in which the other models were preferable, namely inference time. Multiple forward passes ($\sim 10$) was required for our model, while the others only need one.

The most obvious failure mode of the models was when they made small predictions, clearly smaller than the intended lesion. This failure mode inspired us to implement relative area postprocessing. The postprocessing improved metric scores, but its use is limited and probably specific to the LIDC dataset. 

The LIDC dataset is very narrow, and too easy when centrally cropped. The level of ambiguity is quite limited and mostly described by whether or not a ground truth is negative or positive with small deviations across positive masks. We tried to alleviate this with random cropping, but a more diverse dataset seems to be required to further advance the models. It is very costly to segment, hence why datasets with multiple ground truths per image are hard to come by. Previously the Prob. UNet paper~\cite{prob_unet} also tried using synthetic perturbations of data to generate multiple ground truths, but this approach also has limitations.

We tested some diffusion techniques which yielded no improvement. Namely self-conditioning~\cite{bit_diffusion} which has seen impressive results in image generation. This seemed to have no effect, probably because the diffusion reverse process is much more static in our task for than for image generation. Classifier free guidance has also been very successful in other domains, but this actually decreased performance since it consistently caused our our model to over-segment.

\section{Conclusion}
Our model achieved SOTA in uncertain image segmentation, showing that diffusion models do indeed beat the older generative VAE-based models. We explored the space of diffusion model design, by trying different noise schedules, loss weightings and prediction types. We found using a noise schedule with input scaling gave significant boost in performance. We found $v$ and $x$-prediction much better than $\epsilon$-prediction. The best loss weightings were those that gave significant weight to the end of the diffusion process, for example the truncated SNR loss weighting. Our ablations demonstrated that the performance gains were not due to newer architectural details.

\section{Acknowledgements}
This work was supported by Danish Data Science Academy, which is funded by the Novo Nordisk Foundation (NNF21SA0069429) and VILLUM FONDEN (40516).

%% file: text/6supplementary.tex
\section{Supplementary Material}
\subsection{Relative Area Postprocessing}
In order to improve performance measures further, we introduce a postprocessing step called relative area postprocessing. For a set of masks, $\mathcal{M}$, we find the maximum area, $A_\text{max}$ measured in pixels. The relative threshold parameter $r \in [0,1]$ decides how large a mask must be relative to the largest mask to be kept. Formally, the relative area postprocessing is given by
\begin{equation}
    A_\text{thresh} = r A_\text{max} = r \max_{\mathbf{m} \in \mathcal{M}} \sum_{i,j} \mathbf{m}_{i,j} 
\end{equation}
and
\begin{equation}
    \widehat{\mathcal{M}} := \Big\{
    \begin{cases}
        \mathbf{0}, & \text{if } \sum_{i,j} \mathbf{m}_{i,j} < A_\text{thresh}, \\
        \mathbf{m}, & \text{otherwise}, 
    \end{cases}
    \Big\}_{\mathbf{m} \in \mathcal{M}}
\end{equation}
where $\widehat{\mathcal{M}}$ is the set of masks after postprocessing and $\mathbf{0}$ is the empty mask (all zeros). For example, if the largest mask has $A_\text{max}=150$ foreground pixels and $r=0.5$, then all predicted masks with less than $A_\text{thresh}=75$ pixels will be replaced with an empty mask. Note that the postprocessing has no effect if $r=0$. Varying the postprocessing parameter for our model and the AA Gen. Prob. UNet yields a significant performance improvement, but only

\begin{figure}[h]
    \centering
    \includegraphics[width=0.8\textwidth]{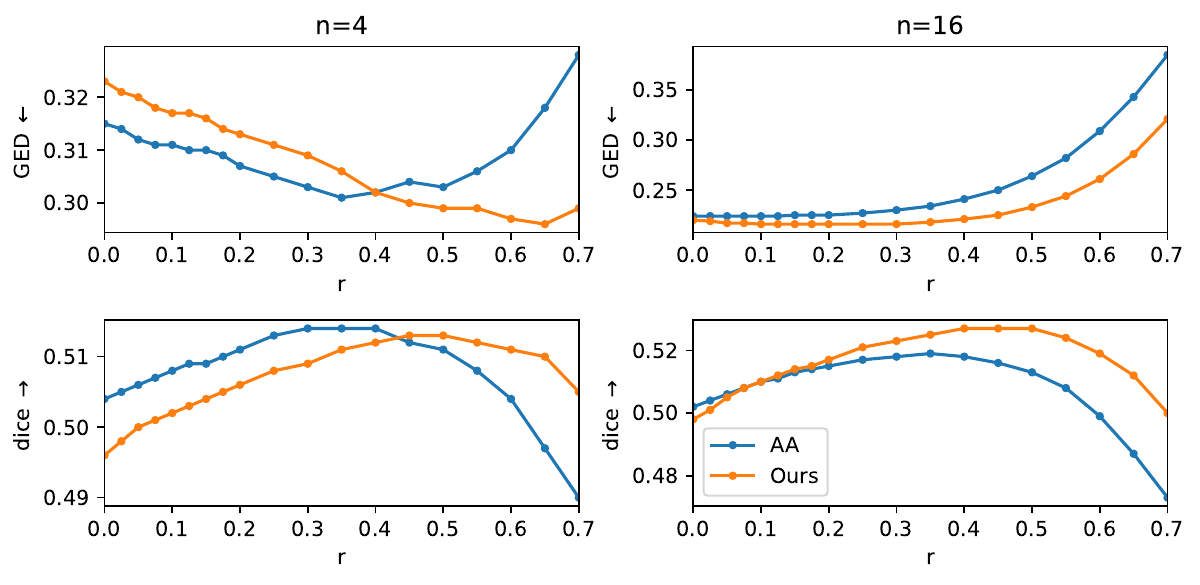}
    \caption{Postprocessing effect measured in GED and IoU as the relative area parameter ($r$) is varied. The number of predictions ($n$) is set to 4 and 16.}
    \label{fig:postprocess}
\end{figure}

The postprocessing (pp) parameter was set to $r=0.5$ and chosen based on the validation set performance of the best current model (AA) and our model as seen in \cref{fig:postprocess}. The GED metric only seems to improve when the number of predictions is relatively low with $n=4$ seeing significant lowering while $n=16$ having almost no effect. Our model benefits slightly more from postprocessing, which could be explained by small noisy patches surviving the denoising process. 

\subsection{Training Details}

Our model was trained with the AdamW~\cite{adamw} optimizer, using pytorch default hyperparameters. The learning rate was set to $10^{-4}$ and the training batch size was 8. Gradient clipping with a maximum normalization value of $1.0$ was used. The models from the ablation study were trained for 50k iterations with 10k iterations of cosine learning rate decay. The final models, which were used in comparisons with others were trained for 400k steps 100k steps of cosine learning rate decay. 

When reproducing the studies and models from \cite{phi_seg,prob_unet,gen_prob_unet,gen_prob_unet2}, we generally made sure to use their reported settings, or to do a basic search for good values. Models were then trained for enough time to reach full convergence. In the case of Phi-Seg~\cite{phi_seg} this was 500k steps and the Gen. Prob. Unets~\cite{prob_unet,gen_prob_unet,gen_prob_unet2} needed 200k steps. Training was somewhat unstable for these models, and we therefore used the best checkpoint as measured by validation set GED. Our model simply used the last checkpoint after all training iterations were completed.

\subsection{Architectural details}

The UNet architectural details are specified in this section. The basic sequential structure of a residual block (ResBlock) in our network was 
\begin{enumerate}
    \item Group Normalization with a group size of 32
    \item SiLU activation
    \item Convolutional layer, kernel size $3\times3$
    \item Group Normalization with a group size of 32
    \item SiLU activation
    \item Convolutional layer, kernel size $3\times3$
\end{enumerate}
A residual connected was also added from the neurons before and after these layers. If the number of channels changed ($\texttt{in\_channels}\neq \texttt{out\_channels}$ in the first conv. layer) then a $1\times1$ would be used to match the channel number for the residual connection.

A different number of these ResBlocks are applied to the stack of neurons at each resolution. The channel dimension was also increased at deeper (downscaled) resolutions in the UNet. We ended up using $\texttt{num\_res\_blocks=[1,2,3,4]}$ and $\texttt{channels=[32,32,64,128]}$. Additionally, 4 ResBlocks were added between the encoder and decoder (aka the middle blocks). This means there was 3 downscaling and upscaling operations. An image of size $128\times128$ would thus yield a $16\times16$ neuronal stack at the lowest resolution.

The network used a self-attention mechanism at the lowest resolution after each ResBlock. The attention mechanism was a standard QKV-attention where each downscaled pixel was viewed as a token, and with just 1 attention head. This means there were a total of 12 attention layers (4 encoder layers, 4 middle blocks, 4 decoder layers). We used skip-connections between each ResBlock of the encoder and decoder. To join the neuron stacks, we used concatenation in the channel dimension.